%% file: iclr2021_conference.tex
\documentclass{article} 
\usepackage{iclr2021_conference,times}

\input{math_commands.tex}

\usepackage{hyperref}
\usepackage{url}

\usepackage{graphicx}
\usepackage{multirow}

\title{Adversarially Robust Segmentation Models Learn Perceptually-aligned Gradients}

\author{Pedro Sandoval-Segura\\
Department of Computer Science\\
University of Maryland\\
College Park, MD \\
\texttt{psando@cs.umd.edu} \\
}

\iclrfinalcopy 
\begin{document}

\maketitle

\begin{abstract}
The effects of adversarial training on semantic segmentation networks has not been thoroughly explored. While previous work has shown that adversarially-trained image classifiers can be used to perform image synthesis, we have yet to understand how best to leverage an adversarially-trained segmentation network to do the same. Using a simple optimizer, we demonstrate that adversarially-trained semantic segmentation networks can be used to perform image inpainting and generation. Our experiments demonstrate that adversarially-trained segmentation networks are more robust and indeed exhibit perceptually-aligned gradients which help in producing plausible image inpaintings. We seek to place additional weight behind the hypothesis that adversarially robust models exhibit gradients that are more perceptually-aligned with human vision. Through image synthesis, we argue that perceptually-aligned gradients promote a better understanding of a neural network's learned representations and aid in making neural networks more interpretable. 

\end{abstract}

\section{Introduction}
\label{section-introduction}

Suppose you are seated at your computer with Photoshop open. On your screen is an image of a dog. If you were asked to make it look like a cat, what kinds of changes would you make? Maybe you would change the shape of the head and ears, or maybe you would add whiskers. But it is unlikely you would add zebra stripes to the fur or extend its neck like a giraffe. You certainly would not add seemingly random noise to the image. 

As humans, our visual system has learned what kinds of features are associated with different classes of objects, and so changing the class of a particular object depends on changing its pertinent features. Ask a deep neural network (DNN) to turn an image of a dog into a cat, and the network will make uninterpretable changes to a collection of pixels such that the resulting image will be a grainy-looking dog, but it will classify it as a cat.

Nevertheless, in a variety of computer vision tasks, DNNs have become the state-of-the-art method. Whether it be classification (\citep{krizhevsky2012imagenet}; \citep{simonyan2014very}), segmentation \citep{long2015fully}, or detection (\citep{ren2015faster}; \citep{redmon2015you}), the performance of DNNs has improved as data, model size, and available compute has increased. But despite their surging popularity and performance, DNNs have an Achilles heel: a weakness to adversarial attacks\footnote{Adversarial attacks are not the \textit{only} weakness of DNNs. Some may argue that model interpretability, the need for large amounts of data, and expensive training costs, among other things, are also huge challenges.}. In the context of image classification, the goal of an \textit{adversarial attack} is to produce an adversarial example that is sufficiently similar to a sample from the dataset, but which causes the DNN to misclassify. Put simply, the decision boundaries learned by even state-of-the-art models appear to be brittle, easily fooled by minor perturbations to the input. 

A variety of defenses against adversarial attacks have been proposed. One of the most popular techniques is known as \textit{adversarial training}, which involves augmenting minibatches with adversarial examples during training \citep{madry2017towards}. A model is said to be robust against adversarial attacks if it correctly labels adversarially perturbed images.

Recent works have shown that robust models are capable of impressive image synthesis tasks due to their learned representations \citep{santurkar2019image}. More specifically, robust classifiers learn gradients that are more perceptually-aligned and which allow a simple optimization procedure to produce realistic image synthesis. By exploring how to perform image synthesis using semantic segmentation networks, we demonstrate that perceptually-aligned gradients can also be learned by segmentation networks. We believe that image synthesis using simple optimization is a good way to understand why a DNN produced a given output. 

\section{Adversarial Attacks}

Adversarial attacks are designed to produce adversarial examples. An adversarial example is a sample of data which has been modified slightly to cause erroneous output from the machine learning model. In some cases, the modifications to the original sample are so minor that even humans are unable to differentiate the original sample from the adversarial one. Of course, if most humans consider two samples of data to be identical, a machine learning model should too. Nevertheless, machine learning models remain vulnerable.

Adversarial attacks can be classified as white-box or black-box based on the amount of knowledge the adversary is assumed to possess. Additionally, attacks can be classified as targeted or untargeted depending on the desired outcome. In a targeted attack, the attack is considered successful only if the adversarial example is misclassified as the target class chosen by the adversary. In an untargeted attack, the attack is successful if the adversarial example is misclassified, regardless of the new class. In the following attack overviews, we will assume an image classification scenario for simplicity, but the attacks can extend to other vision tasks too.

\subsection{White-box Methods}
In a white-box setting, the adversary is assumed to have full knowledge of the model: type, architecture, parameters, and trainable weights. Suppose we have a set of class labels $C$. Given a classifier $f: [0,1]^{d} \rightarrow \R^{|C|}$ and a clean image $x \in [0,1]^{d}$ with ground truth label $y_{\textrm{true}} \in \R^{|C|}$, we can devise a variety of adversarial attacks by using the classifier. While we focus our attention on classic attacks, note that other white-box attacks exist.

\textbf{L-BFGS.} One of the earliest methods to find adversarial examples was proposed by \citet{szegedy2013intriguing} and used L-BFGS to solve the following optimization problem:

\begin{equation} 
\delta^{*} = \argmin_{\delta} \|\delta\|_2 \text{\quad such that \quad} f(x+ \delta) = y_{\textrm{target}} \text{\quad and \quad} x + \delta \in [0,1]^d
\end{equation}

In this case, it is assumed that $y_{\textrm{target}} \in \R^{|C|}$ and that $y_{\textrm{target}} \neq y_{\textrm{true}}$. If the optimization is successful, the adversarial image $x_{\textrm{adv}} = x + \delta^{*}$ is created by adding the minimal perturbation $\delta^{*}$. 

There are a couple drawbacks to this method: the optimization procedure could be slow and the attack could be defended against by degrading the image quality \citep{kurakin2018adversarial}.

\textbf{Fast Gradient Sign Method (FGSM).} \citet{goodfellow2015explaining} argued that the primary cause of DNNs vulnerability to adversarial perturbations was their linear nature. Under this assumption, the authors devised an attack to that was both fast and effective. An adversarial example can be constructed using FGSM in the following way:

\begin{equation}
x_{\textrm{adv}} = x + \epsilon \cdot \sign(\nabla_{x}\mathcal{L}(x, y_{\textrm{true}};\theta))
\end{equation}


\noindent
where $0 < \epsilon < 1$ is a perturbation budget, $\mathcal{L}(x, y; \theta)$ is the loss function used to train the neural network with parameters $\theta$, and where $\sign(\cdot)$ computes the element-wise sign. The perturbation budget controls how large the perturbation can be. FGSM works by linearizing the loss function in an $\ell_{\infty}$ neighborhood of the clean image.

\textbf{Iterative FGSM (I-FGSM).} \citet{kurakin2017adversarial} extended the FGSM attack by applying it multiple times with small step-size, and clipping pixel values of the intermediate image to ensure it remained in an $\epsilon$-neighborhood of the original image. This method is also referred to as the Basic Iterative Method (BIM). Their procedure can be described as follows:

\begin{equation}
x_{\textrm{adv}}^{0} = x, \quad x_{\textrm{adv}}^{N+1} = \textrm{clip}_{x, \epsilon}(x_{\textrm{adv}}^{N} + \alpha \cdot \sign(\nabla_{x}\mathcal{L}(x_{\textrm{adv}}^{N}, y_{\textrm{true}};\theta)))
\end{equation}

\noindent
where $0 < \alpha < 1$ is the step-size for each FGSM step, and where $\textrm{clip}_{x, \epsilon}(\cdot)$ clips its input to remain within the $\epsilon$-ball. Being an extension of FGSM, the procedure is written assuming we are linearizing the loss in an $\ell_{\infty}$ neighborhood of the clean image but, with minor modifications, we could have also chosen to constrain the perturbation to a different $\ell_p$ neighborhood around the clean image. It is possible to turn this attack into a targeted one by replacing $y_{\textrm{true}}$ with $y_{\textrm{target}}$ and moving in the direction opposite of the signed gradient. 

\textbf{Projected Gradient Descent (PGD).} \citet{madry2017towards} showed that I-FGSM can be greatly improved by starting at a random point within the $\epsilon$ norm ball. The only difference between this method and I-FGSM is that we set $x_{\textrm{adv}}^{0} = x + \epsilon \cdot \mathcal{U}(-1,1)$. Because \citet{madry2017towards} consider PGD to be ``the strongest attack utilizing the local first-order information about the network,'' we make primary use of this attack for the adversarial training performed in Section~\ref{section-experiments}.

\subsection{Black-box Methods}
In a black-box setting, the adversary is assumed to have limited or no knowledge of the target model. If the adversary is allowed to provide input and observe the output of the model, the setting is referred to as \textit{black-box with probing}. In general, attack procedures in a black-box setting take advantage of the transferability of adversarial examples \citep{szegedy2013intriguing}, where an adversarial example created for one model is likely to be effective against a different model. This means that the adversary can employ a surrogate model to create an adversarial example, and deploy the adversarial input to a different target model. Research has shown that adversarial examples can transfer despite differences in model architecture \citep{szegedy2013intriguing}, training data \citep{papernot2016transferability}, and even vision task \citep{lu2019enhancing}.

\textbf{Universal Adversarial Perturbations (UAP).} Not only are adversarial examples transferable across model settings, but they have also been shown to transfer across samples of data. \citet{moosavi2017universal} demonstrated the existence of universal perturbations: image-agnostic perturbation vectors that can be applied to any image and which still cause labels estimated by the neural network to change with high probability. While the original algorithm to craft UAPs can be classified as a white-box attack, \citet{mopuri2018generalizable} developed a data-free UAP attack which does not require access to the target model. 

\subsection{Why do adversarial examples exist?}

Many theories to explain the existence of adversarial examples have been proposed. Some work has focused on theoretical models (\citep{schmidt2018adversarially}; \citep{bubeck2019adversarial}) while others have focused on the dimensionality of image data (\citep{gilmer2018adversarial}; \citep{shafahi2018adversarial}), but it is likely there is not a single answer to this question. 

\citet{ilyas2019adversarial} proposed a particularly interesting argument: adversarial vulnerability arises as a direct result of a model's sensitivity to well-generalizing features in the data. They argued that, in a model's quest to maximize accuracy, \textit{any} available signal is taken advantage of. To corroborate this hypothesis, they created a ``non-robust dataset'' of adversarial examples by perturbing clean images from CIFAR-10 toward a target class and relabeling the sample according to the target class. For example, an image from this non-robust dataset might look like a horse (contain robust features of the horse class), but be labeled as a dog (since the sample contains a perturbation toward the dog class). In this way, every adversarial image had robust features of the original class, but non-robust features of the target class. They trained a classifier on this non-robust dataset and found that the standard classification accuracy could be maintained despite the completely mislabeled data, providing evidence that models use non-robust features to make predictions.

Additionally, to demonstrate that adversarial vulnerability arises in part because of the dataset, \citet{ilyas2019adversarial} construct a ``robust dataset'' and show that training on this dataset yields models with good adversarial robustness. While their construction of the robust dataset involves the use of an existing model which is robust to adversarial perturbations, their finding demonstrates that adversarial vulnerability is not tied to the standard training framework. In other words, with the right kind of data, it is possible to train models resistant to adversarial attacks using only standard training. 

\section{Defenses against Adversarial Attacks}

A model is said to be \textit{adversarially robust} if it is largely unaffected by adversarial attacks. Being that most perceptible adversarial examples seem to contain high-frequency noise patterns, it makes sense that early attempts to defend against adversarial attacks were detection methods focused around compression or reducing the precision of the input \citep{xu2017feature}. When the adversary is not aware of the detection technique (a black-box attack setting), defenses like these have proved effective. In fact, the winner of the NIPS 2017 defenses competition was a DNN-based denoiser \citep{kurakin2017adversarial}. But when the adversary is aware of the defense (a white-box setting), defenses of this kind have been shown to fail \citep{he2017adversarial}. If an adversary is aware of the detector being used, they could optimize an adversarial example to simultaneously fool both the detector and the target model.

Another category of defenses are those that use ``gradient masking.'' These defense methods are designed to make a model's gradient information useless by changing the model to make it non-differentiable, creating zero gradients in most places, or having gradients which do not point in the direction of the decision boundary. Since most white-box attacks require computing gradients, making gradient information unusable or inaccessible is a reasonable defense. However, because gradient masking defenses do little to change the model's decision boundaries, these defenses can't protect against black-box transfer attacks. 

To date, the most popular defense against adversarial attacks is known as adversarial training. We dedicate most our focus to adversarial training, as that is the training technique used in our experiments.

\subsection{Adversarial Training}

Adversarial training is an effective defense technique which involves training on adversarial examples. To introduce the concept using the optimization view, we begin by formalizing the training objective for standard classification. Suppose we have a distribution $\mathcal{D}$ over pairs of examples $x \in [0,1]^d$ with corresponding labels $y \in \R^{|C|}$. Let $\mathcal{L}(x, y; \theta)$ be a loss function for the neural network with parameters $\theta$. We can write our training objective as:

\begin{equation}
\min_{\theta} \E_{(x,y) \sim \mathcal{D}}\left[\mathcal{L}(x, y; \theta) \right]
\end{equation}

\noindent
which means we are minimizing the risk. While this objective works well for standard classification, the resulting classifiers are vulnerable to adversarial examples. So, it makes sense to augment the objective by incorporating an adversary. More specifically, we will assume that for every sample $x$, the adversary has a set of allowable perturbations $\Delta(x)$. In this work, we focus on $\ell_{\infty}$ bounded attacks. As expected, the adversary's objective will be to maximize the loss of the neural network by choosing a suitable perturbation $\delta \in \Delta(x)$ to add to the clean data. In this way, the adversarial training objective becomes:

\begin{equation}
\min_{\theta} \E_{(x,y) \sim \mathcal{D}}\left[ \max_{\delta \in \Delta(x)} \mathcal{L}(x + \delta, y; \theta)\right]    
\end{equation}

\noindent
which is a saddle point problem. The inner maximization problem captures the goal of an adversary which seeks to  cause erroneous model output, while the outer minimization problem captures the goal of minimizing the adversarial loss. The theoretical value of this optimization problem also presents a quantitative measure of robustness. If a model were able to achieve zero risk for this adversarial problem, we could say the model is perfectly robust to all adversarial attacks in the adversary's threat model. While $\ell_{\infty}$ bounded attacks are a common threat model, developing more realistic threat models remains an open problem.

In classic adversarial training, at every iteration, the entire batch of clean images is perturbed to craft a new batch of adversarial examples using a white-box attack like FGSM or PGD. The white-box attack is executed on current model parameters which, after the forward and backward pass, will we be updated by the optimizer. This adversarial min-max game is played out at every batch through the training set. One common variant of adversarial training involves \textit{substituting} samples in the clean batch with some ratio of adversarial examples. In either case, \citet{madry2017towards} demonstrate that using a PGD adversary yields robustness against \textit{all} first-order adversaries. Due to this result and others, a PGD adversary was used for adversarially training the models in Section~\ref{section-experiments}.

\section{Perceptually-aligned Gradients}

Surprisingly, adversarial training not only serves as a defense against adversarial attacks, but it also aids DNNs in learning salient data characteristics. Other benefits of adversarial training include regularization-free feature visualization and approximately invertible representations \citep{engstrom2019adversarial}. Previous work also demonstrated that adversarially training helps models learn representations that align well with human perception (\citep{tsipras2018robustness}; \citep{zhang2019interpreting}). It turns out those representations are enough to be able to perform impressive image synthesis tasks.

One particularly exciting work making use of this property is that of \citet{santurkar2019image}, where a single adversarially robust classifier was used for sophisticated image synthesis tasks like generation, inpainting, image-to-image translation, super-resolution, and sketch-to-image. While their results were not meant to compete with popular task-specific synthesis techniques, their work served to illustrate how a simple toolkit consisting of an adversarially robust model and a simple optimization method could be leveraged to solve challenging synthesis problems. For example, recall the hypothetical problem from Section~\ref{section-introduction} of converting an image of a dog into a cat. Using an adversarially-trained classifier, \citet{santurkar2019image} took an image of a dog and used PGD to minimize the loss of the cat target class. They found that the resulting image begins to look like a cat. Their experiments indicate that robust classifiers exhibit perceptually-aligned gradients.

The relationship between adversarial robustness and perceptually-aligned gradients has received recent attention. \citet{aggarwal2020benefits} demonstrated that adversarially training models with a weak adversary results in little to no robustness against adversarial attacks, but that models still exhibit perceptually-aligned gradients. Their results suggested that a model with perceptually-aligned gradients does not guarantee adversarial robustness. \citet{kaur2019perceptually} found that using randomized smoothing resulted in classifiers which exhibited perceptually-aligned gradients and gains in adversarial robustness. Unfortunately, it was unclear by how much their technique underperformed compared to adversarial training. Nevertheless, their results suggested that there may be other learning methods for training models to possess perceptually-aligned gradients. \citet{salman2020adversarially} found that adversarially robust models, while less accurate, performed better than standard-trained models when used for transfer learning. They suggested that perceptually-aligned gradients may be a desirable prior from the perspective of transfer learning. In general, while perceptually-aligned gradients do not guarantee robustness to adversarial attacks, they are a desirable model property to obtain. The exact principles behind the relationship between adversarial robustness and perceptually-aligned gradients remains unclear.

\section{Experiments}
\label{section-experiments}

Our experiments are inspired by \citet{santurkar2019image} where they used a single robust classifier to perform a variety of image synthesis tasks. However, rather than training a classifier, we adversarially train segmentation networks for the purpose of performing image inpainting. In contrast to image classification datasets, where a single label is provided for an image which may contain a variety of different objects, semantic segmentation datasets provide precise ground truth segmentation masks for different object classes present in an image. Thus, we hypothesized that if a segmentation network could be adversarially-trained, its gradients would be more perceptually-aligned than those of a robust classifier, and lead to better inpainting.

\subsection{Models}
\label{subsection-models}

Due to their simplicity, we employ fully convolutional networks for semantic segmenation (FCN) models with a variety of different backbones \citep{long2015fully}. The four models we consider in our experiments are an off-the-shelf pretrained FCN-ResNet50 (\textbf{Pre}), a robust backbone FCN-ResNet50 (\textbf{RB}), an adversarially-trained FCN-ResNet50 (\textbf{AT-50}), and an adversarially-trained FCN-ResNet101 (\textbf{AT-101}). All models are implemented in PyTorch \citep{pytorch}.

We train our models on a subset of the COCO dataset \citep{lin2014microsoft}, on 21 semantic classes (including the background class) that are present in the PASCAL VOC dataset \citep{everinghamVOC2010}. The \textbf{Pre} model was trained on COCO 2017 and is readily available through the PyTorch library. The \textbf{RB} model uses a ResNet-50 backbone which was adversarially-trained on ImageNet with an $\ell_{2}$-norm PGD adversary with $\epsilon=3.0$. The parameters for this robust backbone are downloaded from \citet{robustness}, but the classifier parameters of the \textbf{RB} model are subsequently fine-tuned on VOC 2012 for 20 epochs. Both \textbf{AT-50} and \textbf{AT-101} were adversarially-trained using an $\ell_{\infty}$ PGD adversary with $\epsilon = 0.03$, $3$ steps, and a step size of $0.01$. The \textbf{AT-50} model was adversarially-trained on COCO 2017 for 9 epochs. The \textbf{AT-101} model was adversarially-trained on COCO 2017 for 3 epochs. Due to computational constraints, \textbf{AT-50} and \textbf{AT-101} could not be trained to completion in a reasonable amount of time. 

To evaluate our models, we use both global pixel accuracy and the mean of class-wise intersection over union (mIoU). All evaluations shown in Table~\ref{table:evaluation} are performed on the COCO 2017 validation set, which consists of $5,000$ images. We also perform an evaluation of robust accuracy, where all samples from the validation set are perturbed by the same $\ell_{\infty}$ PGD adversary used to train the \textbf{AT-50} and \textbf{AT-101} models. As expected, \textbf{Pre} obtains the lowest global accuracy and mIoU when under attack by the PGD adversary. Pretrained models are expected to be highly vulnerable to white-box attacks. By using a robust backbone, \textbf{RB} is able to achieve both higher accuracy and mIoU. Lastly, adversarial training, used to train \textbf{AT-50} and \textbf{AT-101}, helps in obtaining even higher global accuracies. Note that it is likely \textbf{AT-101} underperforms \textbf{AT-50} because it was not trained for the same number of epochs.

\begin{table}[]
\centering
\begin{tabular}{|c|c|c|c|c|}
\hline
\multirow{2}{*}{Model} & \multicolumn{2}{c|}{Standard} & \multicolumn{2}{c|}{Robust} \\ \cline{2-5} 
                       & Global Accuracy ($\uparrow$)     & mIoU ($\uparrow$)   & Global Accuracy ($\uparrow$)    & mIoU ($\uparrow$)   \\ \hline
Pre                    & \textbf{91.4}                & \textbf{60.5}    & 20.5                   & 2.0       \\ \hline
RB                     & 82.1                & 19.0    & 68.4                   & 7.9       \\ \hline
AT-50                  & 88.2                & 32.2    & \textbf{80.4}                   & \textbf{15.5}       \\ \hline
AT-101                 & 82.0                & 6.8     & 74.5                   & 4.5       \\ \hline
\end{tabular}
\caption{Standard and Robust evaluations of our models using the COCO 2017 validation set.}
\label{table:evaluation}
\end{table}

\subsection{Inpainting using a Robust Segmentation Model}
\label{subsection-inpainting}

Semantic segmentation is the task of providing a class label for every pixel in an image. Image inpainting is the task of recovering missing pixels in a manner that is perceptually plausible given the rest of the image (\citep{bertalmio2000inpaint}; \citep{hays2007scene}). While a semantic segmentation model is not meant to perform image inpainting, we use inpainting success as one proxy for evaluating the perceptual-alignment of gradients. 

Suppose we are dealing with $C$ semantic classes. Given an image $x \in [0,1]^d$ with a corrupted region corresponding to the binary mask $m \in \{0,1\}^d$, and the ground truth segmentation mask $y \in \{0, 1, \dots, C \}^d$ of the uncorrupted image, the goal is to use a segmentation network to restore the missing pixels. To do this, we optimize the image by minimizing the loss of the true segmentation while penalizing changes to the uncorrupted region, as in \citet{santurkar2019image}. The final inpainted image $x_{I}$ is found by solving:

\begin{equation} \label{eq:inpainting}
x_{I} = \argmin_{x'} \mathcal{L}(x', y; \theta) + \lambda \|(x - x') \odot (1-m) \|_{2}
\end{equation}

where $\mathcal{L}(x, y; \theta)$ is the cross-entropy loss, $\odot$ denotes the Hadamard product, and $\lambda$ is a constant. In practice, we set all pixels in the corrupted region to the per-channel average pixel value of the entire image. To solve this optimization problem, we use PGD. Since the optimization procedure relies on gradient information to minimize Equation~\ref{eq:inpainting}, if a model is able to inpaint the corrupted region in a plausible way, this would imply the model's gradients capture patterns that are perceptually-aligned with human vision.

\begin{figure}[t]
\begin{center}
\includegraphics[width=\textwidth]{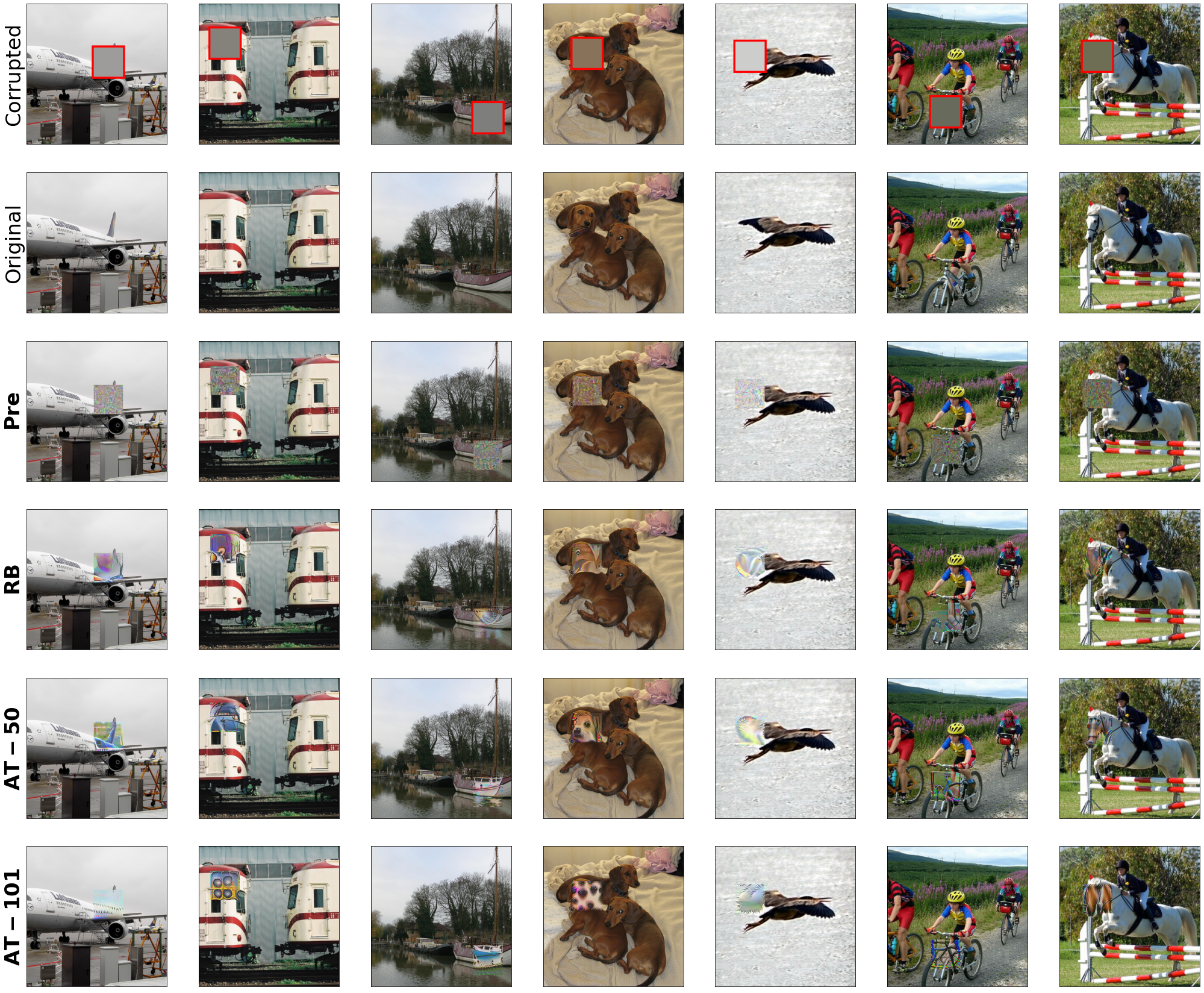}
\end{center}
\caption{Seven images from the VOC 2012 validation set are corrupted within a $100 \times 100$ pixel region. Each model is tasked with optimizing the corrupted region to maximize the score of the ground truth segmentation mask. The adversarially-trained models, \textbf{AT-50} and \textbf{AT-101}, indeed inpaint the corrupted region with plausible pixels in most cases. In contrast, the pretrained model, \textbf{Pre}, only inpaints high-frequency patterns.}
\label{fig:inpainting-results}
\end{figure}

In Figure~\ref{fig:inpainting-results}, we show sample inpaintings obtained by optimizing (\ref{eq:inpainting}) for different FCN models described in Section~\ref{subsection-models}. The resulting inpaintings demonstrate that adversarially-trained models, \textbf{AT-50} and \textbf{AT-101}, are able to inpaint perceptually plausible colors and shapes most of the time. On the other hand, \textbf{Pre} appears to fill the corrupted region with high-frequency patterns. Consider the first column of Figure~\ref{fig:inpainting-results}, which depicts a Lufthansa aircraft on the tarmac. Note that \textbf{AT-50} is able to inpaint a rudder and complete the cabin with a consistent shape. More impressively, \textbf{AT-101} fills the corrupted region with what seem to be additional windows and a white cabin exterior.

Our inpainting results also demonstrate that using a robust backbone in an FCN network, as in \textbf{RB}, is not enough to perform image synthesis tasks like inpainting. While a robust backbone is not enough to have perceptually-aligned gradients, it certainly seems to help. As seen in Table~\ref{table:evaluation} and Figure~\ref{fig:inpainting-results}, both robust metrics and inpainting results for \textbf{RB} are better than the baseline pretrained model.

\subsubsection{Sensitivity to location of Corrupted Region}

\begin{figure}[t]
\begin{center}
\includegraphics[width=\textwidth]{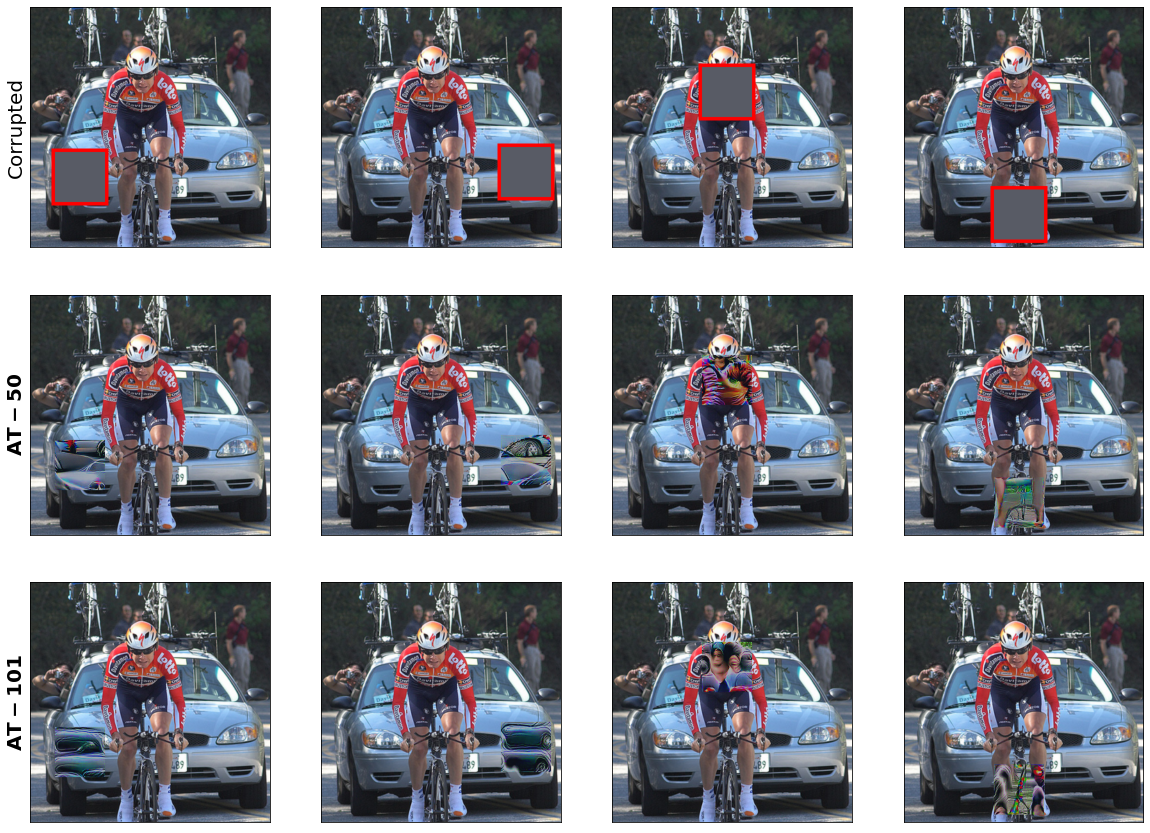}
\end{center}
\caption{Inpainting appears to be sensitive to the location of the corrupted region. From left to right, the columns of images represent inpainting a left headlight, a right headlight, a human face, and feet. Notice how \textbf{AT-50} can inpaint a plausible right headlight, but is unable to produce a plausible and symmetric left headlight.}
\label{fig:sensitivity-results}
\end{figure}

Images of cars are often vertically symmetric when viewed from the front or rear-end. To test whether adversarially-trained segmentation networks capture these characteristics, we hand-picked an image with a car from the VOC 2012 validation set and corrupted the pixel regions corresponding to the left and right headlights. Our inpainting results are shown in  Figure~\ref{fig:sensitivity-results}. Here, we see that neither \textbf{AT-50} nor \textbf{AT-101} inpaint symmetric headlights. \textbf{AT-50} inpaints a great approximation of a right headlight, but is unable to inpaint a plausible left one. On the other hand, \textbf{AT-101} seems to come close to having some symmetry, but its inpainting of the headlights do not resemble headlights. 

The adversarially-trained segmentation networks make similar errors when impainting the biker's face and feet in Figure~\ref{fig:sensitivity-results}. Both \textbf{AT-50} and \textbf{AT-101} are unable to inpaint the subtle patterns required for a human face. However, when it comes to inpainting feet, both networks appear to match the biker's skin color in the inpainted legs. Inpaintings for \textbf{Pre} and \textbf{RB} are not shown in Figure~\ref{fig:sensitivity-results} since the quality of their inpaintings are not reasonable enough to warrant attention.

Recall that the optimization procedure of Equation~\ref{eq:inpainting} defines no incentive to inpaint reasonable object features; rather, PGD uses gradient information from the segmentation network to perturb the corrupted region in manner that maximizes the score of the ground truth mask. Perceptually-aligned gradients present in \textbf{AT-50} and \textbf{AT-101} are the sole reason why inpainting a particlar region results in perceptually reasonable patterns.

\subsection{Image Generation using a Robust Segmentation Model}

Another proxy we use to evaluate the perceptual-alignment of gradients is image generation. The formulation of the image generation problem presented here is a generalization of inpainting, where the optimization procedure is allowed to modify any and all pixels. Rather than be constrained by the rest of the image (as in image inpainting), the optimizer is free to make any necessary changes so that the resulting generated image, when passed to a segmentation network, will induce a mask prediction that is close to the ground truth segmentation mask.

Given an image $x \in [0,1]^d$ with ground truth segmentation mask $y \in \{0, 1, \dots, C \}^d$, the goal is to use a segmentation network to make changes to any pixel to maximize the extent which the predicted mask matches the true mask. To do this, we optimize an image by solely minimizing the loss of the predicted mask given the true segmentation, as in \citet{santurkar2019image}. The final generated image $x_{G}$ is found by solving:

\begin{equation} \label{eq:generation}
x_{G} = \argmin_{x'} \mathcal{L}(x', y; \theta)
\end{equation}

\begin{figure}[t]
\begin{center}
\includegraphics[width=\textwidth]{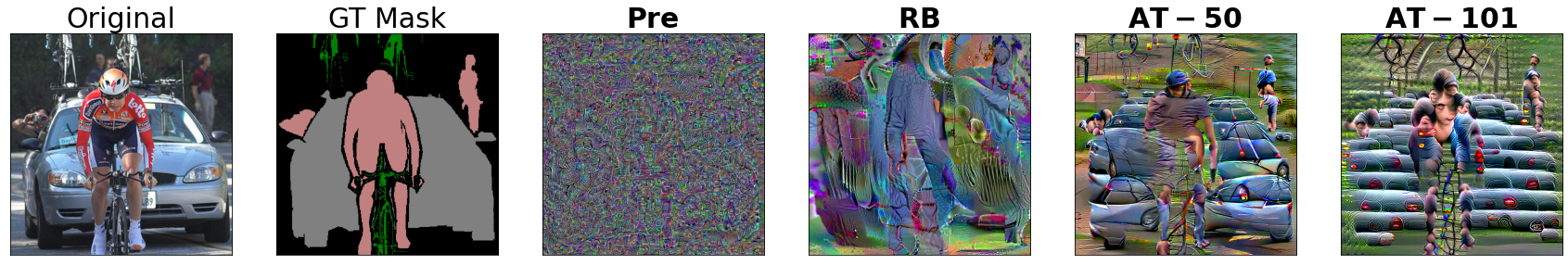}
\end{center}
\caption{Image generation given the ground truth segmentation mask (GT Mask). The models do not see the original image. \textbf{AT-50} and \textbf{AT-101} optimize toward pseudo-shapes and occasionally consistent colors. }
\label{fig:generation-results}
\end{figure}

As in Section~\ref{subsection-inpainting}, we use PGD to solve this optimization problem. The optimization procedure begins from an image where all pixels are the per-channel average pixel value of the original image. The generated images are shown in Figure~\ref{fig:generation-results}.

Gradient information from \textbf{Pre} seems to optimize toward high-frequency patterns, as usual. However, there does appear to be faint clusters which outline the region where the human should be located. The gradients from \textbf{AT-50} are used to generate an image that is particularly interesting. The generated image contains a sea of quasi-cars (with windows and wheels) in regions where the cars should be. The image generated using \textbf{AT-50} also seems to have human legs and part of a human arm. It is difficult to interpret exactly what might be going on in the image generated using \textbf{AT-101}, as it is more abstract, but there certainly are some human faces in the region where a human should be. In regions where a car should be, there is a multitude of what might be car tail lights.

\subsection{Toward Learning Perceptually-aligned gradients}

Perceptually-aligned gradients show up in models that have been adversarially-trained, but there may be other ways of training models to possess these kinds of gradients. We posit that it may be possible to ``train the gradient'' along with minimizing empirical risk. Solutions to this problem may not be so simple, however, since it is not clear whether training for perceptually-aligned gradients will yield any robustness, and there is evidence that robustness comes at the price of standard accuracy \citep{tsipras2018robustness}. Training the gradient could consist of penalizing gradients which do not point in a direction that could change the current sample from being perceptually similar to a different class. We expect to focus on these challenges for future work.

\section{Conclusion}

By adversarially training segmentation models, we provide further evidence that robust models exhibit perceptually-aligned gradients. We take advantage of these learned representations to perform image inpainting and generation. By comparing to a pretrained model, we illustrate how standard training is insufficient in providing robustness or perceptually-aligned gradients. We also incorporate a segmentation model with a robust backbone to demonstrate that if part of the model is robust, gains in robust accuracy occur, but they do not compare to adversarially training the entire model. 

Unfortunately, image inpainting and generation using a semantic segmentation network does not produce images that are as realistic as those from \citet{santurkar2019image}, where they use a classifier. Assuming that this discrepancy is not due to the short training time of our models, this is surprising. During training, a segmentation network has access to per-pixel labels through ground truth segmentation masks. In contrast, a classifier has access to a single label for all pixels in the input image. One could expect that this difference in supervision at training time would result in segmentation networks that learn representations of local object features, allowing for better image inpainting or generation. A diametrically opposed argument might claim that a single label for an image, as in classification datasets, allows a classifier to learn global object features, which lead to less noisy gradients. In any case, new methods for adversarially training segmentation models may resolve this difference in synthesis capabilities.

For DNNs to continue their ascendancy in vision, two prominent issues should be addressed: robustness and interpretability.  When gradients are perceptually-aligned, it can be easier for a human to understand why the model made a certain choice. When minor perturbations to the input do not cause wild changes the model's output, a human can be more confident in the model's ability to generalize and operate in the real world. Adversarial training is a popular solution that addresses both of these issues, but more work is necessary to make it practical for use in semantic segmentation networks.

\subsubsection*{Acknowledgments}
We would like to thank Vasu Singla, David W. Jacobs, and Kfir Aberman for helpful discussions throughout the period of this work.

\bibliography{iclr2021_conference}
\bibliographystyle{iclr2021_conference}


\end{document}

%% file: math_commands.tex
\usepackage{amsmath,amsfonts,bm}

\def\eqref#1{equation~\ref{#1}}

\def\1{\bm{1}}

\DeclareMathAlphabet{\mathsfit}{\encodingdefault}{\sfdefault}{m}{sl}
\SetMathAlphabet{\mathsfit}{bold}{\encodingdefault}{\sfdefault}{bx}{n}

\newcommand{\E}{\mathbb{E}}

\newcommand{\R}{\mathbb{R}}

\DeclareMathOperator*{\argmin}{arg\,min}

\DeclareMathOperator{\sign}{sign}